\documentclass[conference]{IEEEtran}
\IEEEoverridecommandlockouts
\usepackage{cite}
\usepackage{amsmath,amssymb,amsfonts}
\usepackage{algorithmic}
\usepackage{graphicx}
\usepackage{textcomp}
\usepackage{xcolor}
\usepackage{booktabs}    
\usepackage{epsfig}
\usepackage{multicol}

\usepackage{caption}
\usepackage{subcaption}

\usepackage[breaklinks]{hyperref}
\hypersetup{
    colorlinks=true,
    citecolor=blue,
    linkcolor =	red,
    filecolor=magenta,      
    urlcolor=cyan,
}

\usepackage{multirow,tabularx,capt-of}
\def\BibTeX{{\rm B\kern-.05em{\sc i\kern-.025em b}\kern-.08em
    T\kern-.1667em\lower.7ex\hbox{E}\kern-.125emX}}
\begin{document}

\title{Graph Neural Network based Child Activity
Recognition}

\author{
\IEEEauthorblockN{Sanka Mohottala, Pradeepa Samarasinghe, Dharshana Kasthurirathna} 
\IEEEauthorblockA{\textit{Faculty of Computing,} \\
\textit{Sri Lanka Institute of }\\
\textit{Information Technology}\\
Sri Lanka\\
sanka.m@sliit.lk, pradeepa.s@sliit.lk, dharshana.k@sliit.lk}
\and
\IEEEauthorblockN{Charith Abhayaratne} 
\IEEEauthorblockA{\textit{Department of Electronic and} \\
\textit{Electrical Engineering,}\\
\textit{University of Sheffield}\\
United Kingdom \\
c.abhayaratne@sheffield.ac.uk}

}

\maketitle

\begin{abstract}
This paper presents an implementation on child activity recognition (CAR) with a graph convolution network (GCN) based deep learning model since prior implementations in this domain have been dominated by CNN, LSTM and other methods despite the superior performance of GCN. To the best of our knowledge, we are the first to use a GCN model in child activity recognition domain. In overcoming the challenges of having small size publicly available child action datasets, several learning methods such as feature extraction, fine-tuning and curriculum learning were implemented to improve the model performance. Inspired by the contradicting claims made on the use of transfer learning in CAR, we conducted a detailed implementation and analysis on transfer learning together with a study on negative transfer learning effect on CAR as it hasn't been addressed previously. 

As the principal contribution, we were able to develop a  ST-GCN based CAR model which, despite the small size of the dataset, obtained around 50\% accuracy on  vanilla implementations. With feature extraction and fine tuning methods, accuracy was improved by 20\%-30\% with the highest accuracy  being 82.24\%. 
Furthermore, the results provided on activity datasets empirically demonstrate that with careful selection of pre-train model datasets through methods such as curriculum learning could enhance the accuracy levels. Finally, we provide preliminary evidence on possible frame rate effect on the accuracy of CAR models, a direction future research can explore.



\end{abstract}



\begin{IEEEkeywords}
Child Action Recognition, Graph Neural Networks, Transfer Learning, Feature Extraction, Fine Tuning, Curriculum Learning  
\end{IEEEkeywords}


\section{Introduction}
\label{section:intro}

Human Action Recognition (HAR) methods based on skeleton data have been widely investigated and received considerable attention due to high accuracy  achieved on benchmark datasets such as NTU RGB+D~\cite{shahroudy2016ntu} and Kinetics~\cite{carreira2017quo}.  Recently most of the methods that achieve the state-of-the-art (SOTA) accuracy on benchmark datasets have been based on graph convolutional network (GCN) deep learning methods~\cite{yan2018spatial,shi2019two,shi2020skeleton}. This accuracy increment can be attributed to many factors such as view invariance, robustness to occlusion and segmentation of the skeleton. 

Spatial Temporal Graph Convolutional Network (ST-GCN)~\cite{yan2018spatial} architecture was the first to utilize graph neural networks for skeleton-based human activity recognition and had achieved the SOTA results outperforming CNN and LSTM based models~\cite{yan2018spatial,peng2021stgcnnew}. Furthermore, it has been used with small and distinct datasets in different scenarios including fall detection~\cite{Keskes2021VisionBasedFD,zheng2019fall}, hand sign detection~\cite{amorim2019,li2019hand} and others~\cite{10.1145/3341162.3345581,tsai2021}.

Child activity recognition (CAR) has important applications in video game development~\cite{10.1145/3424636.3426909}, early detection of autism~\cite{zunino2018video,zhang_application_2021}, safety monitoring~\cite{goto2013}, object-play behavior assessment~\cite{Westeyn2012} and many others. As most of the HAR models are based on adult datasets and as previous studies in motion style transfer~\cite{10.1145/3424636.3426909} and pose estimation~\cite{sciortino_estimation_2017} have shown that due to differences in size, anatomy, and motion, these models can't be used for CAR, it is required to develop robust and well generalized CAR models.

Early CAR models as well as small dataset based CAR models have utilized signal processing methods along with classical machine learning methods~\cite{rehg2013decoding,tsiami2018}. Deep Learning (DL) methods have  been used lately with image based, skeleton based and wearable sensor based approaches through CNN~\cite{huang2015human,silva2021skeleton}, LSTM~\cite{dechemi2021,Hadfield2018} and other methods~\cite{suzuki2012activity,efthymiou2018multi}. As GCN based models have not been applied in the past on CAR, we direct this research to fill that gap.

Though there are claims made on the availability of public datasets for CAR~\cite{lemaignan_pinsoro_2018,sun2019,Rajagopalan2013,marinoiu2018}, few of those were not  accessible due to non-availability and few were not capturing the whole body. Most of the existing DL research on CAR have not used the limited public datasets but has demonstrated the results on private datasets. Addressing this research gap, we have considered two standard Kinect camera based child activity datasets named Kinder-gator~\cite{Aloba2018kindergator}, Child-Whole-Body-Gesture (CWBG)~\cite{Vatavu2019CFBG} and a standard motion capture system based dataset named kinder-gator 2.0~\cite{dong_yuzhu_2020_4079507} and we present the first evaluation of CWBG dataset on a SOTA DL model.

Though transfer learning (TL) can be used to overcome the limited dataset challenge in public datasets,~\cite{efthymiou2018multi,suzuki2012activity,sciortino_estimation_2017} suggest that TL from adult to child is a challenging task resulting in undesirable performance.
While there is a TL implementation with ST-GCN architecture~\cite{Keskes2021VisionBasedFD} in HAR, it lacks a detailed study of different TL approaches and TL performance on ST-GCN architecture.


Based on the literature study and the gaps identified, in this paper we provide the following main contributions, 
\begin{itemize}
\item To the best of our knowledge, this is the first implementation of a GCN based model for child activity recognition.
\item We provide the first benchmark results on a publicly available child activity dataset using ST-GCN model.
\item Our research shows that comparable results can be achieved through transfer learning with ST-GCN model and demonstrate performance and comparative analysis with different learning approaches.
\item A pre-train dataset selection method is introduced in this research to improve transfer learning and reduce negative transfer learning based on curriculum learning concept.

\end{itemize}

The rest of the paper is organized as follows. 
The models and methods used in the ST-GCN based implementations are discussed in Section~\ref{section:methodology}. 
Pre-processing process used with each dataset and the experimental setups used in each case are discussed in Section~\ref{section:experiments}. The performance of learning methods and the best results achieved by each method on different protocols are discussed in  Section~\ref{section:results,discussion} while Section~\ref{section:conclusion} concludes with future research directions.

\section{Methodology}
\label{section:methodology}

Based on the ST-GCN original paper~\cite{yan2018spatial} and on the official ST-GCN PyTorch implementation~\cite{mmskeleton2019}, a TensorFlow based model was implemented and further experiments were carried out based on this model with NTU RGB+D dataset used for the pre-training models.

Two approaches were applied to quantitatively evaluate the ST-GCN model on the available child activity datasets:
\begin{itemize}
    \item Standard Deep Learning: Train the ST-GCN model on the child activity datasets directly and do the hyperparameter tuning to attain a suitable model,
    \item Transfer Learning: Use a pre-trained ST-GCN model to leverage the learnt feature representations.
\end{itemize}

\subsection{Learning Methods}
\label{Sec:learning_methods}
\begin{figure}[tbp]
    \centering
    \includegraphics[width=8cm]{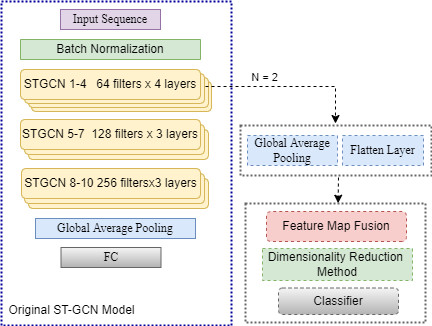}
    \caption{ST-GCN model and feature extraction pipeline}
    \label{fig:STGCN_model}
\end{figure}

\subsubsection{Standard Deep Learning}
Under this approach, ST-GCN model was directly trained with child activity datasets. Due to the small size of these datasets, the original ST-GCN model with 256 channels in each of 10 ST-GCN layers could appear to contain excessive capacity. Since excessive capacity result in over-fitting~\cite{wang2016machine}, model tuning was also attempted with  different number of filters as detailed in Section~\ref{subsection: SDL}.

\subsubsection{Transfer Learning}
Lack of acceptable scale datasets generally results in poor performance of deep learning models. As all the openly available child activity datasets are small in size, developing improved learning methods was essential. Several TL approaches such as fine tuning~\cite{2018,5288526,tan2018survey}, and feature extraction~\cite{zhuang2020comprehensive,zhang2019transfer,sharif2014cnn_feature_extraction} have been used in the subsequent implementations to improve the performance of the model.

\paragraph{Fine Tuning Method}
\label{Para:fine_tuning_methods}
Fine tuning of the pre-trained ST-GCN model was done using several methods as detailed in TL literature~\cite{yosinski2014transferable,zhang2019recent}.  

\begin{enumerate}
  \item Frozen layer approach - Fine tuning $n$ top ST-GCN layers, where $1\leq n < 10$. 
  \item Hybrid approach - In the original ST-GCN model in Fig.~\ref{fig:STGCN_model}, $n$ top ST-GCN layers were randomly initialized where $1\leq n \leq 10$.  
  \begin{itemize}
     \item Hybrid-Frozen : Combines feature extraction and standard deep learning together. Bottom $10-n$ ST-GCN layers were kept frozen.
     \item Hybrid-FineTuned : Combines fine tuning and standard deep learning together. Bottom $10-n$ ST-GCN layers were fine tuned.
   \end{itemize}
  \item Propagation approach - Fine tuning all  ST-GCN layers.  
\end{enumerate}
 
Since ST-GCN contains only a single layer classifier, initial experiments were done with a single randomly initialized dense layer. Experiments were later done with the FC classifiers with up to 4 dense layers. A dropout layer was used as a regularization method when overfitting occurred. Hyper-parameter tuning was also applied to get the best accuracy.

\paragraph{Feature Extraction Method}
Inspired by~\cite{awais_can_2020,yu_feature_extract_stratified_2017}, the feature representations were extracted from feature maps of the ST-GCN model (Fig.~\ref{fig:STGCN_model}). The original ST-GCN model was then supplemented with either a flattening layer or a global average pooling (GAP) layer as the intermediate layer between the ST-GCN model and the classifier. Fusion of feature maps  was done with consecutive maps from two  and three layers of ST-GCN.  Enhancing this approach further, dimensionality reduction techniques such as principle component analysis (PCA), truncatedSVD were employed.   
Experiments were done with support vector machine (SVM), logistic regression as well as feed forward neural network (FFNN) as the classifier.

\subsubsection{Curriculum Learning}\label{section:CL}
To improve the efficiency of both Standard Deep Learning and TL, Curriculum Learning (CL)~\cite{bengio2009curriculum}     was applied, where the model was trained starting from easy samples and gradually exposing to more challenging samples. While there are several variants in implementing CL, our implementation contained Scoring and Pacing functions as in~\cite{hacohen2019power}:   
\begin{itemize}
    \item Scoring Function: gives the probability of level of difficulty of each sample, calculated by training the model with 10\% of epochs.
    \item Pacing Function: decides the number of samples used in each epoch enabling us to add the samples based on the descending order of the values generated by the Scoring Function. This was implemented as a step function with varying number of steps.
\end{itemize}

\subsection{Activity Datasets}
\label{section:Activity Datasets}
 A preliminary research was done on the existing child activity datasets and found that all three publicly available datasets, Kinder-Gator~\cite{Aloba2018kindergator}, Kinder-Gator 2.0~\cite{DongYuzhu2020kindergator} and CWBG dataset~\cite{Vatavu2019CFBG} were depth sensor based and were available only in skeleton modality. In order to overcome the challenge in limited data sets, we used existing adult skeleton mode activity data sets, Kinect v2 based NTU-120 and NTU-60 for building the pre-trained models. NTU-120 was taken as the main dataset, which  contains 850 videos with an average number of frames  ranging from 76 to 300. Actions are performed by 106 participants above 10 years of age.
 
 To improve the TL and to reduce the negative TL effect~\cite{wang2019characterizing}, several subsets of NTU  were identified.

\subsubsection{Large Scale Datasets}
\label{SubSec:NTU-51}
\begin{itemize}
     \item NTU-120: The full NTU RGB+D-120 dataset was used with a different dataset splitting method as detailed in Section~\ref{SubSec:Data Pre} than the one proposed in~\cite{2020ntu120}. With this approach, we were able to remove any potential bias resulting from an unbalanced data distribution.
     \item NTU-60: We used the full NTU RGB+D dataset along with 11 interaction classes.
     \item NTU-51: Out of the datasets of NTU-60, in order to minimize ambiguities in activity identification, 9 interaction classes were removed resulting in 49 single action classes and 2 interaction classes.
   \end{itemize}
   
\subsubsection{Curriculum Learning Inspired Datasets}
\label{SubSec:CurriculumDatasets}
While there are other methods~\cite{minmax_minghui2019,tute_sibgrapi_2017}, we used a simplified CL inspired approach to select better classes. Best classes are chosen by analysing the confusion matrices of NTU-60/120 based STGCN models. 
\begin{itemize}
    \item NTU-44: Out of those sorted, 44 classes were selected. To reduce ambiguities introduced from spatial and temporal symmetrical classes, we kept only one such class in this subset.
    \item NTU-22: Enhancing the approach taken for NTU-44 further, a more discriminative dataset of 22 classes was introduced. Classes were selected by analysing the NTU-44 confusion matrix.
\end{itemize}
\subsubsection{NTU Frame Rate Adjusted (NTU-FRA)}
Since the CWBG dataset is recorded with an approximate 10FPS frame rate, we introduced three down-sampled data sets.

\begin{itemize}
    \item NTU-44-FRA : Selected each of the 3rd frame in every sequence in NTU-44 subset. 
    \item NTU-60-FRA : Selected each of the 3rd frame in every sequence in NTU-60 subset.
    \item NTU-120-FRA : Selected each of the 3rd frame in every sequence in NTU-120 subset.
\end{itemize}

\subsection{Skeleton Structure}
\label{SubSec:skeleton}
As the NTU dataset was created with Kinect v2 and CWBG dataset was created using Kinect v1, output skeleton structure generated in NTU and CWDG are different to each other.  

Transfer learning when applied to convolution based models usually takes a input as a grid structure similar to images. Thus, when the target domain data samples differ in size from source domain, resizing with standard interpolation techniques can be utilized. But with graph structure, structural change has to be taken into consideration and knowledge transferability in GCN based models is still not fully developed~\cite{zhu2021transfer}. Thus we use 20 shared joints with removal of 5 joints from NTU source dataset.

\subsection{Child Dataset Protocols}
The CWBG dataset is released for the study of child gesture elicitation and contains 1312 sequences from 30 children between the ages of 3 - 6 with an equal gender distribution.

\begin{itemize}
    \item CWDG-Full: Contains 15 classes and is used to test the performance of entire dataset.
    \item CWDG-Similar: Contains 10 classes, removing most dissimilar classes. This group contains challenging classes even for a human (Table~\ref{CWBG classes} - column Challenging).
    \item CWDG-Dissimilar: Contains 10 classes including most dicriminative  (Table~\ref{CWBG classes} - column Discriminative).

\end{itemize}

\section{Experiments} \label{section:experiments}

This section discusses the stages followed in conducting
the quantitative analysis including: pre-processing of the datasets, and the experimental settings.

\subsection{Data Pre-Processing}
\label{SubSec:Data Pre}

In the first stage of data pre-processing, noisy data such as empty sequences and pseudo-skeleton sequences were removed. If a class belongs to a human-human interaction, then the action of the person with the most activity movements was selected in both NTU-60 and NTU-120 datasets whereas these ambiguous actions were removed in NTU-51,  NTU-44, and  NTU-22 datasets.

In the next stage, following process was followed for all the data sets.

\begin{itemize}
    \item Fixed the frame size to 300, increasing feature visibility. 
    \item First order information was extracted by translating each skeleton such that the spine joint position in each frame is the origin of coordinate system (i.e., [0,0,0]).
    \item Rotated the skeleton around spine joint such that person is looking towards positive x-axis and spine is parallel to z-axis.
\end{itemize}

In the final stage, dataset was split into train and validation subsets according to the cross subject method~\cite{shahroudy2016ntu}  for NTU-60 and CWDG based datasets. For additional classes other than NTU-60 in NTU-120 based activity datasets, training and validation was split randomly 70\% to 30\%.





\begin{table}[]
\begin{center}
\caption{Challenging and Discriminative Child Whole-Body Gesture (CWBG) Dataset classes}
\label{CWBG classes}
\begin{tabular}{ll}
\hline
\addlinespace[2pt]
Challenging classes & Discriminative classes \\ \addlinespace[2pt]
\hline
\addlinespace[1pt]
Draw a circle       & Hands up               \\
Draw a square       & Crouch                 \\
Draw a flower       & Jump                   \\
Angry like a bear   & Applaud                \\ \hline
\addlinespace[1pt]
\end{tabular}
\end{center}
\end{table}


\subsection{Implementation Details}

\subsubsection{Standard Deep Learning}
Standard deep learning approach was implemented with original ST-GCN model without any changes to the architecture and other hyper-parameters were also kept same except for the learning rate scheduler.\\ 
Further implementations were done to change the model capacity by adjusting the number of filters in each layer but the ratio of filter number between layers was kept at the same value as in original ST-GCN model. New layer filter number was changed by $R$,  $R={\bar{F}_{n}}/{F_{n}}$, where
$n$ refers to ST-GCN layer number ($1\leq n \leq 10$) while $\bar{F}$ refers to new model filters and $F$ to the original model filters. 
\subsubsection{Transfer Learning}
\paragraph{Pre-trained model}
\label{Para:Pre-trained model}
The approach used for implementing ST-GCN in this research differs with the previous implementations in a number of ways. 
\begin{itemize}
\item
Since use of dropout and batch normalization together result in degrading results~\cite{garbin_dropout_2020}, dropout layers were not used as outlined in~\cite{yan2018spatial}.
\item The implemented TensorFlow based ST-GCN uses an equal weighting system for the graph rather than a trainable weight mask as in~\cite{yan2018spatial}, since the mask weights are intrinsic to the dataset used for training. Hence those weights won't be adaptable in TL.    
\end{itemize}

Training was done for 30 epochs, with a batch size of 4, using stochastic gradient descent (SGD) optimizer. The learning rate was initialized to 0.1 and was dropped to 0.01 and further to 0.001 using the piece-wise constant decay scheduler.

\paragraph{Fine Tuning}
To evaluate fine-tuning  approach, experiments were done using the NTU-44 dataset based pre-trained model. In frozen layer approach, FC layer as in Fig.~\ref{fig:STGCN_model} was replaced with a randomly initialized  dense layer. 

For the comparison of fine tuning approach on three main CWBG dataset protocols, implementations were done on all activity dataset based pre-trained models. Hyper-parameter tuning was done on each to achieve the best accuracy keeping other parameters constant.

\paragraph{Feature Extraction}
All feature extraction approach evaluations were done using the NTU-44 dataset with same hyper-parameters as in fine-tuning evaluation. Performance evaluations were done based on the output layer, use of GAP layer, use of dimension reduction methods and the fusion of feature maps.

On implementing the SVM classifier, linear kernel function with L1 regularization constant set to 1 was applied. For the logistic regression implementation, multi-nominal logistic regression classifier was used with L2 regularization constant kept at 1. For the FFNN based classifier, experiments were done with different combinations of layers and nodes and for the evaluation results single layer classifier was used.

While the performance analysis was carried out with all three classifiers, based on the best performance, only FFNN classifier was used for the comparative analysis.

\subsubsection{Curriculum Learning}
Curriculum learning was implemented with feature extraction  and fine tuning methods on the NTU-FRA dataset based pre-trained models as well as with standard deep learning approach. For pacing function, a non-uniform step function was used such that latter steps last longer (i.e., more epochs) than the previous steps. Same batch size and optimizer was used but learning rate was changed to account for the training behavior resulted from step function used as the pacing function.

Frame rate effect on the performance was evaluated by dropping the frames of NTU-44/60/120 datasets to match the CWBG dataset's frame rate. Experiments were done by taking the moving average of joint position vectors with sliding window of 5 time periods and 3 time periods in order to see if the noisy data of NTU had affected the performance of the model. 
\section{Results and Discussion} 
\label{section:results,discussion}

\begin{figure}[t]
\centering
\small
\begin{tabular}{cc}
\includegraphics[width = 0.48\linewidth, scale = 0.9]{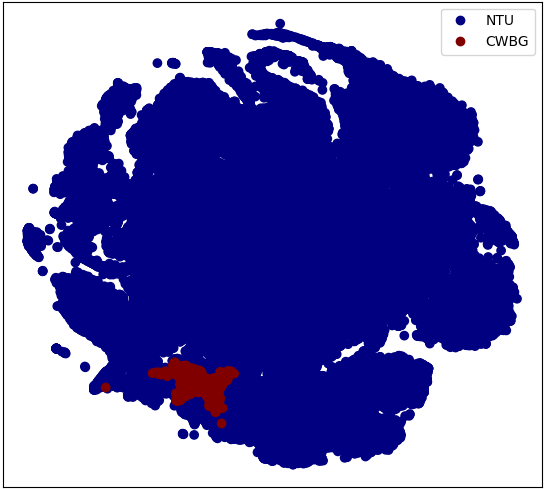} & 
\includegraphics[width = 0.46\linewidth, scale=0.9]{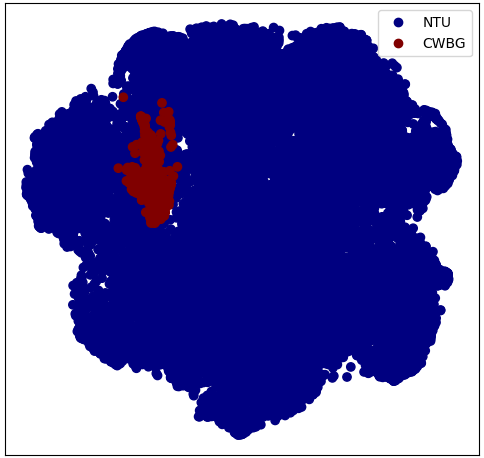} \\
(a) NTU 120 and CWBG  & (b) NTU 44 and CWBG \\
\end{tabular}
\caption{NTU and CWBG dataset visualization}
\label{fig:NTU44_tSNE}
\end{figure}

On the the ST-GCN model implementation, accuracy of 78.7\% was achieved on the NTU RGB+D dataset. While the original authors of~\cite{yan2018spatial} achieves 81.5\% accuracy on the cross-subject protocol, the change of accuracy(-2.8\%) could be explained through the changes done on the ST-GCN architecture as detailed in Section~\ref{Para:Pre-trained model}.

Visualization of NTU-120 and CWBG datasets as well as NTU-44 and CWBG datasets through t-distributed Stochastic Neighbourhood Embedding (t-SNE) (Fig.~\ref{fig:NTU44_tSNE}) shows overlapping distributions, confirming the potential use of TL to increase the accuracy as shown in~\cite{minmax_minghui2019}.

\subsection{Standard Deep Learning}
\label{subsection: SDL}
In the vanilla ST-GCN model implementation with CWBG datasets, when the training accuracy converges to an optimal value, test datasets  reach an accuracy in the range of 40\%-55\% (Table~\ref{T:SDL  accuracy}). Such low accuracy is expected on a limited dataset such as CWBG, yet ST-GCN over performed Random accuracy in multitudes. 
Given that model capacity is extremely high ($\approx$3M parameters) compared to dataset size, over-fitting should happen according to classical machine learning theory~\cite{wang2016machine}. But the results from smaller capacity model implementations (for $R$=1/32, $\approx$5k parameters and for $R$=1/2, $\approx$700k parameters) in table~\ref{T:SDL  accuracy} shows that rather than increasing the accuracy, smaller capacity slightly decreases the accuracy. This suggests that the low accuracy in the vanilla ST-GCN model is not due to over-fitting. These results are consistent with the existing research done on over-parameterized neural networks and the resultant high generalization accuracy~\cite{zhang2021,Zhu2018}. Another implementation done on a higher capacity model (for $R$=2, $\approx$12M parameters) shows this indeed is the case. Hence we can conclude vanilla ST-GCN results are the optimal results on all CWBG protocols.         
\begin{table}[tbp]
\centering
\caption{Standard Deep Learning Top-1 Accuracy}
\label{T:SDL  accuracy}
\resizebox{\columnwidth}{!}{%

\begin{tabular}{llll}
\hline
\addlinespace[2pt]
\textbf{Child Datasets}                    & \textbf{CWDG-Full}  & \textbf{CWDG-Similar} & \textbf{CWDG-Dissimilar} \\
\addlinespace[1pt] \cline{1-1}  \addlinespace[1pt]
                                        
Method              &  &  &   \\
\hline                    \addlinespace[2pt]
 Vanilla ST-GCN              & 41.71      & 47.84      & 53.18      \\
 Vanilla Curriculum Learning & 40.65      & 44.36            & 51.46   \\
 \hline                    \addlinespace[2pt]
 ST-GCN   $R=1/32$            & 39.64      & 42.29      & 47.19      \\
 ST-GCN   $R=1/2$             & 39.12      & 50.16      & 47.57      \\
 ST-GCN   $R=2$             & 27.46      & 44.71      & 57.68      \\
 \hline                    \addlinespace[2pt]
Random                 & 6.67      & 10      & 10      \\
\hline  
\end{tabular}%
}
\end{table}






\begin{table}[tbp]
\centering
\caption{Currriculum Learning based feature extraction results on activity datasets}
\label{T:CL_accuracy}
\resizebox{\columnwidth}{!}{%
\begin{tabular}{llll}
\hline
\addlinespace[2pt]
\textbf{Accuracy} & \textbf{NTU-44-FRA} & \textbf{NTU-60-FRA} & \textbf{NTU-120-FRA}     \\
          \hline
         \addlinespace[2pt]
CWDG-Full & \underline{58.62}         & 53.74        & 25.65         \\
CWDG-Similar  & \underline{63.53}        & 62.14        & 33.56         \\
CWDG-Dissimilar & \underline{79.03}        & 76.78        &44.57  \\
\addlinespace[2pt] \hline  
\end{tabular}%
}
\end{table}

\subsection{Feature Extraction Results}\label{SubSec:FX_results}
\begin{figure*}[tbp]
\begin{subfigure}[b]{.48\linewidth}
\centering
\includegraphics[width=0.8\textwidth]{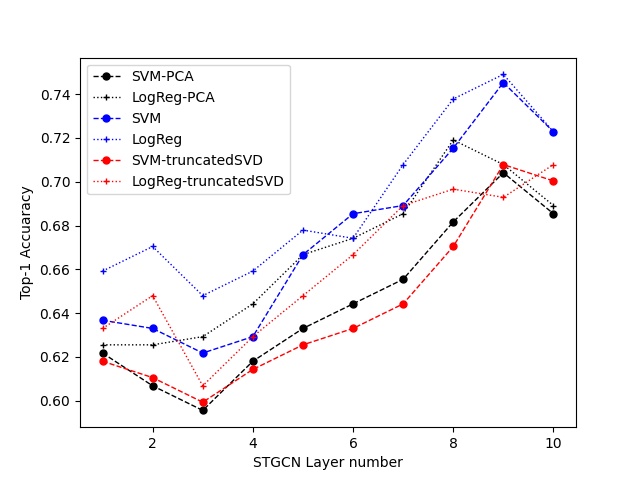}

\caption{Flattening layer approach}
\label{fig:FX_flattening}
\end{subfigure}
\hfill
\begin{subfigure}[b]{.48\linewidth }
\centering
\includegraphics[width=0.8\textwidth]{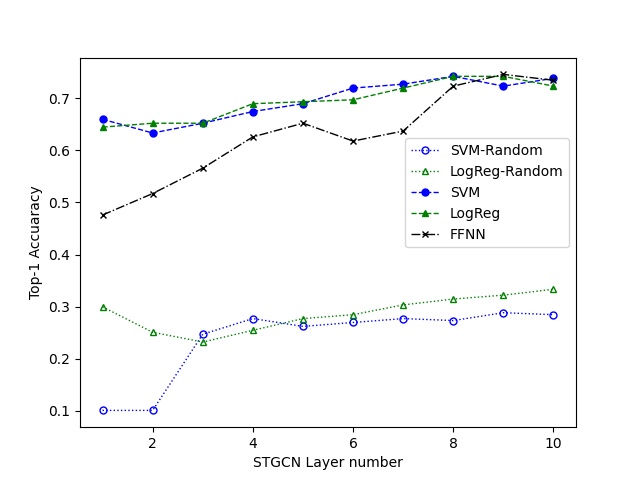}
\caption{Results on ST-GCN layer output}
\label{fig:FX_layerwise}
\end{subfigure}
\hfill
\begin{subfigure}[b]{.48\linewidth}
\centering
\includegraphics[width=0.8\textwidth]{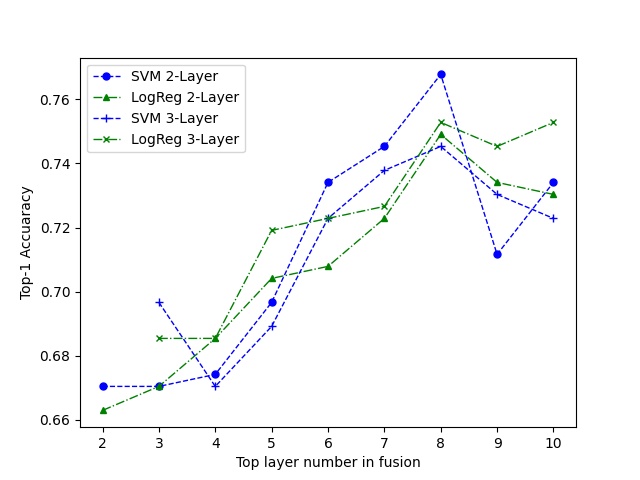}
\caption{Feature map fusion approach}
\label{fig:FX_fusion}
\end{subfigure}
\hfill
\begin{subfigure}[b]{.48\linewidth}
\centering
\includegraphics[width=0.8\textwidth]{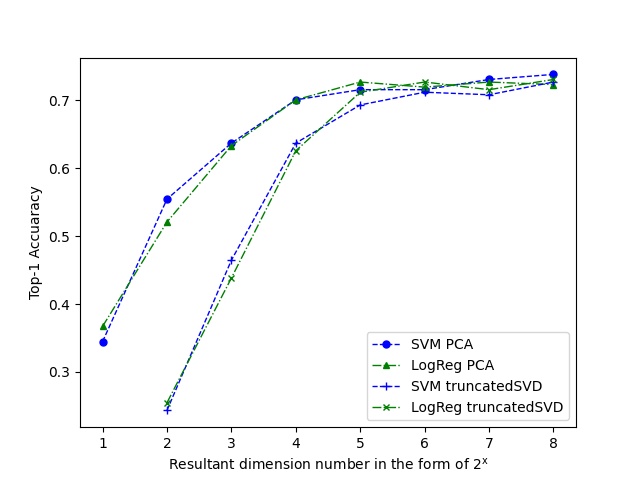}
\caption{Dimensionality reduction approach}
\label{fig:FX_dim_red}
\end{subfigure}
\caption{Feature extraction performance with different configurations}
\end{figure*}

Gradual increase in accuracy as the the output ST-GCN layer number reaches 10 in Fig,~\ref{fig:FX_layerwise} is analogous to the conventional CNN model behaviour in feature extraction~\cite{donahue2014decaf}, thus confirming the existence of a hierarchical feature representation in ST-GCN architecture which is the fundamental requirement for TL applicability. Furthermore, similar increase of accuracy is present in the flatten layer implementation (Fig,~\ref{fig:FX_flattening}) as well as in feature map fusion implementation (Fig,~\ref{fig:FX_fusion}). Compared to the SVM and LogReg classifier feature extraction accuracy (around 65\%-75\%), when an untrained ST-GCN model is used, same classifiers give a lower accuracy (around 25\%-30\%) (Fig.~\ref{fig:FX_layerwise}), thus empirically proving that the resultant feature accuracy is indeed due to the pre-trained model's feature representations rather than the classifier.

SVM and logistic regression results can be considered as the upper bound of feature extraction accuracy. While 1-layer FFNN  performs on par with SVM and LogReg on top ST-GCN layers, its poor performance on bottom ST-GCN layers suggests that feature maps are more linearly separable as layer number increases and this is to be expected in a model that learns hierarchical feature representation~\cite{donahue2014decaf}.

As detailed in Fig~\ref{fig:FX_dim_red} and Figs~\ref{fig:FX_flattening},~\ref{fig:FX_fusion}, even though dimensionality reduction and the use of flattening layers do not increase accuracy, feature map fusion contributes to accuracy improvement slightly.  
\begin{figure}[tbp]
\centering
\captionsetup{justification=centering}
\small
\begin{tabular}{cc}
\includegraphics[width = 0.44\linewidth, scale = 1.2]{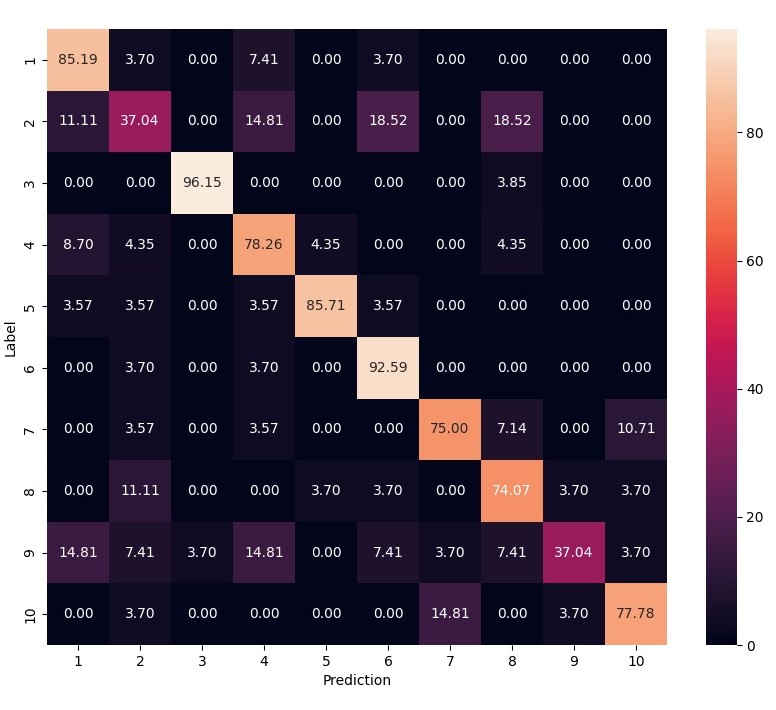} & 
\includegraphics[width = 0.44\linewidth, scale=1.2]{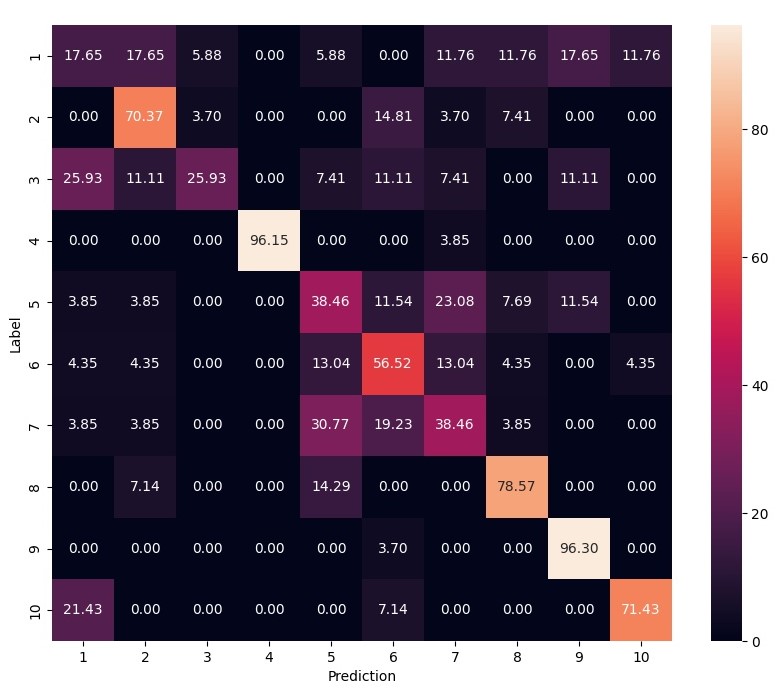} \\
(a) CWBG-Dissimilar protocol  & (b) CWBG-Similar protocol \\
\end{tabular}
\caption{Feature Extraction Performance on NTU-44 based Model}
\label{fig:conf_on_NTU_44}
\end{figure}

\begin{table}[tbp]
\centering
\caption{Feature extraction results on child datasets}
\label{T:FX_results}
\resizebox{\columnwidth}{!}{%
\begin{tabular}{lllllllll}
\hline
\addlinespace[2pt]
\textbf{Accuracy} & \textbf{NTU} & \textbf{NTU} & \textbf{NTU} & \textbf{NTU} & \textbf{NTU} & \textbf{NTU} & \textbf{NTU}  & \textbf{NTU}       \\
         & \textbf{60}  & \textbf{120} & \textbf{51}  & \textbf{44}  & \textbf{22}  & \textbf{44-FRA} & \textbf{60-FRA} & \textbf{120-FRA} \\ \hline
         \addlinespace[2pt]
CWDG-Full    & 47.98     & \underline{52.69}     & 49.48     & 51.26    & 47.41      & 57.77         & 52.85         & \textbf{\underline{59.33}}          \\
CWDG-Similar    & 56.26    &  56.07   & 54.9    & \underline{61.17}    & 52.94    &  \textbf{\underline{67.31}}        & 61.96         & 63.14  \\  
CWDG-Dissimilar    & 70.04    &  73.03   & 73.03    & \underline{74.16}    & 67.79    &  79.10        & 77.15         &  \textbf{\underline{82.24}}  \\  
\addlinespace[2pt] \hline  
\end{tabular}%
}
\end{table}

\subsection{Fine Tuning Results}
In evaluating performance between the two Hybrid  and the Frozen approaches~\ref{Para:fine_tuning_methods}, either of the Hybrid approaches always outperformed the fine tuning approach irrespective of the number of layers in the ST-GCN model (Table~\ref{T:FineTune_layer}). In order to compare the standard fine tuning approach and the Hybrid approach on the child datasets, Hybrid-Frozen and Vanilla Fine-Tuned models were chosen as the models in Table~\ref{T:Fine tuning results}.

\begin{table}[tbp]
\centering
\caption{Layer-wise Fine Tuning Implementation}
\label{T:FineTune_layer}
\resizebox{\columnwidth}{!}{%

\begin{tabular}{lllllll}
\hline
Fine-Tuning  & \multicolumn{6}{c}{ST-GCN layer number}              \\ \cline{2-7} 
        Approaches      & 4     & 5     & 6     & 7     & 8     & 9     \\ \hline
Hybrid-Fine-Tuned                      & 74.91 & 76.78 & 76.4  & 77.53 & 76.03 & 74.91 \\
Hybrid-Frozen                       & 76.78 & 77.9  & 75.66 & 74.53 & 76.03 & 75.28 \\
Vanilla Fine-Tuned                 & 74.96 & 75.28 & 73.41 & 74.53 & 71.54 & 72.28\\
\hline
\end{tabular}%
}
\end{table}
\begin{table}[]

\end{table}

\begin{table}[tbp]
\centering
\caption{Fine-tuning comparison on child datasets}
\label{T:Fine tuning results}
\resizebox{\columnwidth}{!}{%
\begin{tabular}{llllllllll}
\hline
\addlinespace[2pt]
&\textbf{Accuracy} & \textbf{NTU} & \textbf{NTU} & \textbf{NTU} & \textbf{NTU} & \textbf{NTU} & \textbf{NTU} & \textbf{NTU}  & \textbf{NTU}       \\
&         & \textbf{120}  & \textbf{60} & \textbf{51}  & \textbf{44}  & \textbf{22}  & \textbf{44-FRA} & \textbf{60-FRA} & \textbf{120-FRA} \\ \hline
         \addlinespace[2pt]
{\multirow{3}{*}{\rotatebox[origin=c]{90}{\textbf{Vanilla}}}}&CWDG-Full    & \textbf{\underline{58.03} }    & 52.33       & 54.66     & 56.74    & 52.07      & 57.25         & 55.44         & \underline{57.51}          \\
&CWDG-Similar    & \underline{66.27}    &  63.92   & 58.43     & 63.14    & 58.43    &  62.35       & \textbf{\underline{\textcolor{red}{66.67}}}         & 60.78  \\  
&CWDG-Dissimilar    & \underline{76.78}    &  69.66   & 68.54    & 75.28    & 67.04    &  74.91        & 76.4         & \underline{76.78}  \\  
\addlinespace[2pt] \hline         \addlinespace[2pt]
{\multirow{3}{*}{\rotatebox[origin=c]{90}{\textbf{Hybrid}}}}&CWDG-Full    & \textbf{\underline{\textcolor{red}{58.55}}}     & 57.25       & 55.3     & 55.7    & 51.55     & 54.14         & 55.44         & \underline{58.29}  \\
&CWDG-Similar    & \textbf{\underline{65.88}}    &  63.92   & 61.38     & 63.92    & 57.65    &  63.14       & \underline{65.49}         & 65.1  \\  
&CWDG-Dissimilar    & 77.53    &   \underline{78.28}   & 76.4    & 76.4    & 70.79    &  76.03        & \textbf{\underline{\textcolor{red}{78.65}}}         & 78.28  \\
\addlinespace[2pt] \hline
\end{tabular}%
}
\end{table}

\subsection{Curriculum Learning Results}
Vanilla curriculum learning in Standard Deep Learning approach did not result in a significant accuracy improvement (Table~\ref{T:SDL  accuracy}). When CL is applied to feature extraction method (Table~\ref{T:CL_accuracy}), implementations on NTU-44-FRA and NTU-60-FRA results in similar accuracies as in original feature extraction results, yet there is either little improvement or no improvement. 
With NTU-120-FRA, the results are significantly worse than the original feature extraction results. Based on the model loss, there is significant overfitting and with the pacing function we used, it can’t be avoided. These results suggest that even though with CL some improvement can be achieved, it is difficult to tune the model to achieve those results and with more challenging datasets like NTU-120, it is even extremely difficult.

\subsection{Comparative Results}

Out of all the comparative results, the highest accuracy is shown with CWBG-Dissimilar protocol (Tables~\ref{T:SDL  accuracy},~\ref{T:CL_accuracy},~\ref{T:FX_results}, and ~\ref{T:Fine tuning results}).
This result is further corroborated through the generated confusion matrices and 
two such resultant confusion matrices for feature extraction approach are given in Fig.~\ref{fig:conf_on_NTU_44}.\\ 
Strong correlation is evident in the classes with labels 5-7 (Fig.~\ref{fig:conf_on_NTU_44} - (b)) representing the first three classes in 
Table~\ref{CWBG classes} (column Challenging). Action visualization demonstrates label 1 class corresponding to "angry like a bear" doesn't result in considerable skeleton movements, explaining the misclassification. Better accuracies are observed for the more distinct classes of Fig.~\ref{fig:conf_on_NTU_44} - (a), endorsing that ST-GCN model achieves an acceptable accuracy in child action recognition.\\

Detailed feature extraction implementations were done on final layer output without dimensionality reduction and with GAP layer (Table~\ref{T:FX_results}). Best accuracies are achieved by NTU-120-FRA based model for CWBG-dissimilar protocol and CWBG Full protocol while for the CWBG similar, NTU-60-FRA achieves the best accuracy. When 30FPS based activity datasets are compared,  best accuracy for CWBG similar and CWBG dissimilar protocols are achieved by NTU-44, demonstrating our curriculum inspired dataset based model is suitable for TL just as large dataset based models. But the results on NTU-22 shows that when this approach is taken to an extreme with small number of classes, the resultant model’s performance degrades. 

Comparison of fine-tuning models was done with $n=1$ as detailed in~\ref{Para:fine_tuning_methods}. Results on vanilla fine-tuned models in Table~\ref{T:Fine tuning results} suggest that unlike in feature extraction method, largest activity dataset (i.e., NTU-120) based model outperform all the other models including the NTU-44. Even when 30FPS-activity datasets and 10FPS-activity datasets are considered separately, this finding still holds. Comparison results on Hybrid-Frozen models in Table~\ref{T:Fine tuning results} also confirms this interpretation. CWBG-Full protocol and CWBG-Dissimilar protocol perform better on the Hybrid-Frozen approach while CWBG-Similar protocols performs better with the Vanilla Fine-Tuned approach. Contrary to the other implementations~\cite{yosinski2014transferable}, we achieved marginal improvement of Hybrid-Frozen approach over the vanilla Fine-Tuning approach.

\begin{table}[tbp]
\centering
\caption{Benchmark results of CWBG protocols}
\label{T:Benchmark}
\resizebox{\columnwidth}{!}{%
\begin{tabular}{llll}
\hline
\addlinespace[2pt]
Top-1 Accuracy               & CWBG-Full & CWBG-Similar & CWBG-Dissimilar \\ \hline
\addlinespace[2pt]
All activity datasets  & 59.33     & 67.31        & 82.24           \\
30FPS-activity datasets  & 58.55     & 65.88        & 78.28  \\
\addlinespace[2pt] \hline
\end{tabular}
}
\end{table}

To provide the benchmark results, performance of CWBG protocols  accross the learning methods was compared based on two criterion. One considering all the activity dataset based models and the other considering only the 30FPS-activity dataset based models (Table~\ref{T:Benchmark}). While all the 30FPS-activity dataset results were derived from fine-tuning approach, the all activity dataset results were derived from the feature extraction approach. Furthermore, when each NTU-FRA result in Table~\ref{T:FX_results} is compared with it's corresponding 30FPS NTU result, there is a significant increase in accuracy but such increase is not present in Table~\ref{T:Fine tuning results}. These observations provides clear indication of frame rate effect in feature representation and further experiments are needed to evaluate its implications.

\section{Conclusion and Future Works}
\label{section:conclusion}
This paper presents a CAR model based on GCN architecture and a detailed analysis with different learning methods. \\
With the ST-GCN implementation, despite the small size of CWBG dataset, acceptable accuracies of 47.84\% and 57.68\% were achieved on CWBG-Similar and CWBG-Dissimilar, providing the first GCN based CAR model results and demonstrating the applicability of GCN in CAR.

Applying TL approaches such as feature extraction and fine-tuning, we were able to improve these accuracies to 65.88\%  and 78.28\% on 30FPS-activity datasets and to 67.31\% and 82.24\% on 
all activity datasets. These results corroborate TL approaches like feature extraction and fine-tuning can improve accuracy in CAR despite contrary claims made by previous research.


Superior performance of NTU-44 based model over other 30FPS based models in feature extraction gives strong evidence of CL inspired dataset selection process use in TL but NTU-22 performance suggest there needs to be a balance between dataset size and dataset classes. 

These results suggest several future research directions such as  analysis of frame rate effect in TL, detailed study of CL parameter selection and effective methods to achieve a trade off between parameters to avoid negative TL. Moreover, to achieve scalability in practical implementations, a robust approach for TL with different skeleton structures is also essential.

\small{   
\section*{Acknowledgment}
This research was supported by the Accelerating Higher Education Expansion and Development (AHEAD) Operation of the Ministry of Higher Education of Sri Lanka funded by the World Bank (\url{https://ahead.lk/result-area-3/}).
}
\bibliographystyle{IEEEtran}
\bibliography{references}

\end{document}